%% file: main.tex
\documentclass{article} % For LaTeX2e
\usepackage{iclr2027_conference,times}

\input{math_commands.tex}

\usepackage{xurl}
\usepackage{booktabs}
\usepackage{needspace}
\usepackage{graphicx}
\usepackage{wrapfig}
\usepackage{tabularx}
\usepackage{longtable}
\usepackage{colortbl}
\usepackage{placeins}
\usepackage{float}

\usepackage{multirow}
\usepackage{pifont}
\usepackage[hidelinks]{hyperref}

\title{FusionEval: Benchmarking Model Fusion for Large Language Models}

\author{%
  \textbf{Shuo Cai}\textsuperscript{1*},
  \textbf{Yanggan Gu}\textsuperscript{1*},
  \textbf{Yuhang Liu}\textsuperscript{1*},
  \textbf{Su Lu}\textsuperscript{1},
  \textbf{Yuanyi Wang}\textsuperscript{1},
  \textbf{Yang Yu}\textsuperscript{1},\\[3pt]
  \textbf{Zhen Li}\textsuperscript{1},
  \textbf{Zhitian Hou}\textsuperscript{2},
  \textbf{Guanghao Zhu}\textsuperscript{1},
  \textbf{Ming Li}\textsuperscript{1,3\textdagger},
  \textbf{Hongxia Yang}\textsuperscript{1,2,3\textdagger}\\[6pt]
  \textsuperscript{1}The Hong Kong Polytechnic University\quad
  \textsuperscript{2}InfiX.ai\\[2pt]
  \textsuperscript{3}PolyU-Daya Bay Technology and Innovation Research Institute\\[5pt]
  \texttt{shuo1031.cai@connect.polyu.hk}
}

\iclrfinalcopy
\renewcommand{\headrulewidth}{0pt}
\makeatletter
\renewcommand{\@maketitle}{%
  \vbox{\hsize\textwidth
    {\LARGE\scshape\@title\par}
    \vskip 12pt
    {\centering\small\@author\par}
    \vskip 14pt
  }%
}
\makeatother
\hypersetup{
  pdftitle={FusionEval: Benchmarking Model Fusion for Large Language Models},
  pdfauthor={Shuo Cai, Yanggan Gu, Yuhang Liu, Su Lu, Yuanyi Wang, Yang Yu, Zhen Li, Zhitian Hou, Guanghao Zhu, Ming Li, Hongxia Yang}
}

\definecolor{FusionEvalYes}{rgb}{0.18,0.49,0.31}
\definecolor{FusionEvalNo}{rgb}{0.74,0.29,0.29}
\newcommand{\cmark}{\textcolor{FusionEvalYes}{\ding{51}}}
\newcommand{\xmark}{\textcolor{FusionEvalNo}{\ding{55}}}
\newcommand{\FusionEvalTableStyle}{%
  \centering
  \footnotesize
  \setlength{\tabcolsep}{2.5pt}%
  \renewcommand{\arraystretch}{1.05}%
}

\usepackage{capt-of}
\begin{document}

\maketitle
\begingroup
\renewcommand{\thefootnote}{}
\footnotetext{\textsuperscript{*}Equal contribution.\quad \textsuperscript{\textdagger}Corresponding authors.}
\endgroup
\suppressfloats[t] % Keep top floats from preceding the title on the first page.

\begin{abstract}
Model fusion aims to integrate the capabilities of multi-expert models into a single model. Existing evaluations offer limited coverage of reasoning with long chains of thought and tool use, while strong average performance can mask substantial losses in individual expert capabilities. We introduce \textsc{FusionEval}, a benchmark that provides controlled expert pools, integrates fusion methods with and without additional training, and offers a unified evaluation suite. The pools contain domain experts independently trained from Qwen3 and Llama base models across math, code, science, instruction following, and tool use. The suite evaluates capability retention and computational cost on eight public benchmarks under comparable data access rules, reporting normalized average and worst performance, runtime, and peak GPU memory usage. Extensive experiments show that training-free methods preserve reasoning performance but substantially degrade tool use at both model scales. In contrast, MOPD preserves tool use and achieves a more balanced capability profile, but requires additional data preparation, inference, and training. Jointly retaining reasoning, instruction following, and tool use at low computational cost remains a challenge. We hope \textsc{FusionEval} supports future research on efficient model fusion that jointly preserves expert capabilities.
Our code is available at \url{https://github.com/Baicaihaochi/FusionEval_Benchmark}.
\end{abstract}

% Let the overview float naturally after the expanded author block.
\begin{figure}[!ht]
  %\vspace{-10pt} % Reduce the gap above this figure
  \centering
  \includegraphics[
    width=\linewidth,
    %trim={0pt 0pt 0pt 6pt},
    clip
  ]{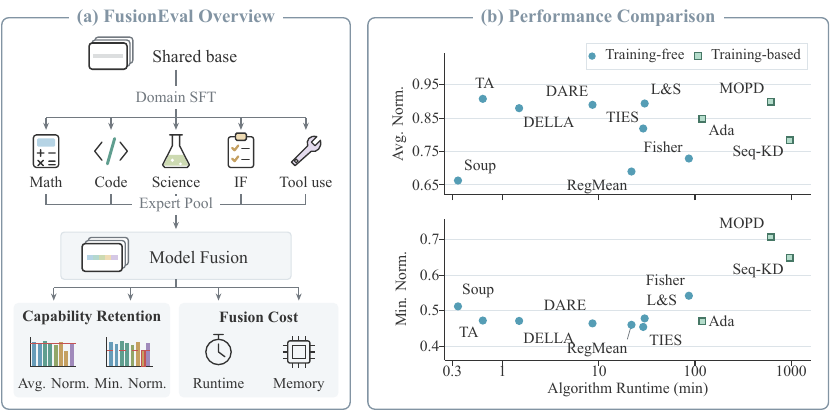}
  \caption{Overview of \textsc{FusionEval}. (a) Domain SFT of a shared base yields experts in five domains. Fusion is evaluated by capability retention, runtime, and peak GPU memory. (b) Qwen3-4B performance versus runtime (minutes, log scale). Avg.\ Norm.\ and Min.\ Norm.\ are the mean and minimum of benchmark scores normalized by the corresponding experts.}
  \label{fig:framework}
\end{figure}

% Comparison table follows the overview in the float order.
\begin{table}[t]
  \caption{Coverage of model fusion evaluations. Training-free and Training-based indicate support for fusion methods without and with gradient updates, respectively. Large model denotes evaluation of LLMs with at least 7B parameters. Reasoning covers math, code, or science tasks. Continual Fusion evaluates fusion as experts arrive sequentially. Tool use evaluates external tool calls.}  \label{tab:benchmark-comparison}
  \FusionEvalTableStyle
  \scriptsize
  \begin{tabularx}{\linewidth}{@{}l>{\hsize=1.00\hsize\centering\arraybackslash}X>{\hsize=1.15\hsize\centering\arraybackslash}X>{\hsize=.95\hsize\centering\arraybackslash}X>{\hsize=.85\hsize\centering\arraybackslash}X>{\hsize=1.30\hsize\centering\arraybackslash}X>{\hsize=.75\hsize\centering\arraybackslash}X@{}}
    \toprule
    \textbf{Evaluation} &
    \makebox[\hsize][c]{\textbf{Training-free}} &
    \makebox[\hsize][c]{\textbf{Training-based}} &
    \makebox[\hsize][c]{\textbf{Large model}} &
    \makebox[\hsize][c]{\textbf{Reasoning}} &
    \makebox[\hsize][c]{\textbf{Continual Fusion}} &
    \makebox[\hsize][c]{\textbf{Tool use}} \\
    \midrule
    Compositional Eval~\citep{tam2024realistic} & \cmark & \cmark & \xmark & \xmark & \xmark & \xmark \\
    Model-GLUE~\citep{zhao2024modelglue} & \cmark & \xmark & \cmark & \cmark & \xmark & \xmark \\
    FusionBench~\citep{tang2024fusionbench} & \cmark & \cmark & \xmark & \cmark & \cmark & \xmark \\
    Merging at Scale~\citep{yadav2025matters} & \cmark & \xmark & \cmark & \xmark & \xmark & \xmark \\
    MergeBench~\citep{he2025mergebench} & \cmark & \cmark & \cmark & \cmark & \xmark & \xmark \\
    \midrule
    \textsc{FusionEval} & \cmark & \cmark & \cmark & \cmark & \cmark & \cmark \\
    \bottomrule
  \end{tabularx}
\end{table}

\section{Introduction}

Model fusion integrates the knowledge and abilities of independently trained models into a single deployable model, reducing the need to deploy and maintain multiple source models~\citep{wortsman2022modelsoups,yadav2023ties,jin2023regmean,ma2026mopd}.
The growing ecosystem of open large language models (LLMs) offers an increasingly diverse pool of reusable models, while fusion methods span weight combination, representation alignment, and multi-teacher distillation~\citep{cai2026fusion}.
Model fusion benchmarks provide a common setting for comparing fusion effectiveness and computational cost~\citep{tang2024fusionbench,he2025mergebench}.

Existing evaluations differ in their coverage of fusion methods, expert models, and capabilities, as summarized in Table~\ref{tab:benchmark-comparison}.
Merging at Scale examines the effects of model size, base model quality, and expert count, but focuses on fusion methods without additional training~\citep{yadav2025matters}.
FusionBench supports methods both with and without additional training~\citep{tang2024fusionbench}, while compositional evaluations study skill combination through tasks such as question answering, summarization, and natural language inference~\citep{tam2024realistic}.
Beyond conventional NLP tasks, FusionBench and Model-GLUE include mathematical reasoning~\citep{tang2024fusionbench,zhao2024modelglue}, and MergeBench standardizes expert training and evaluation across domains including math and code, reporting capability retention and runtime~\citep{he2025mergebench}.
Despite this progress, these evaluations provide limited coverage of experts trained on long chain-of-thought (CoT) responses and of tool use, leaving the joint retention of reasoning, instruction following, and tool use insufficiently tested.
A complementary evaluation setting is therefore needed to examine whether these capabilities can be preserved simultaneously and at what computational cost.

To address these limitations, we introduce \textsc{FusionEval}, a benchmark that provides controlled expert pools, integrates fusion methods with and without additional training, and offers a unified evaluation suite for capability retention and computational cost.
As shown in Figure~\ref{fig:framework}, we use open models from the Qwen~\citep{yang2025qwen3} and Llama~\citep{meta2024llama32} families as bases and select five domains: math, code, science reasoning, instruction following, and tool use.
We select datasets for each domain (Table~\ref{tab:training-data}) and independently apply supervised fine-tuning (SFT) to each base model using the corresponding data, producing a separate pool of five domain experts for each base checkpoint.
Unlike M2RL, which obtains domain experts through reinforcement learning from a shared model trained with mixed SFT data~\citep{wang2026m2rl}, we train each expert directly from the base model through SFT.
This setting reflects the practical scenario of training experts separately for different domains and tests whether fusion retains each expert's capabilities.
We compare fusion methods using the same expert pools, consistent data access rules, and eight public benchmarks.
The evaluation suite normalizes each benchmark score by that of the corresponding domain expert and reports the mean and minimum across benchmarks to assess overall retention and the weakest retained performance.
It also reports fusion runtime and peak GPU memory.

Our experiments show that several fusion methods retain much of the math and code performance of experts trained on long CoT responses, but retention is uneven across reasoning, instruction following, and tool use: high average scores can mask severe losses in tool use.
In contrast, multi-teacher on-policy distillation (MOPD) achieves the highest minimum normalized benchmark score among the evaluated methods at both tested Qwen3 scales, yielding a more balanced capability profile.
However, it requires additional data preparation, inference, and training.
We hope \textsc{FusionEval} encourages the development of efficient model fusion methods that jointly preserve expert capabilities.

\section{Related Work}
\label{sec:related-work}

\paragraph{Model Fusion Methods.}
Model fusion combines the capabilities of multiple models through parameter operations or distillation of model outputs.
Training-free fusion starts with weight averaging and task-vector addition, as in Model Soup (Soup) and Task Arithmetic (TA)~\citep{wortsman2022modelsoups,ilharco2023taskarithmetic}.
To reduce interference between expert updates, TIES resolves sign conflicts, while DARE, DELLA, and Dataless Localize-and-Stitch (L\&S) control which updates are combined through random dropping or magnitude-based selection~\citep{yadav2023ties,yu2024dare,deep2024della,he2025localize}.
Other training-free methods use data statistics to guide parameter combination: Fisher Merging (Fisher) estimates parameter importance, RegMean and LOT-Merging (LOT) solve for weights using layer-input statistics, and FeatCal calibrates fused representations through closed-form updates~\citep{matena2022fisher,jin2023regmean,sun2025towardsminimizing,gu2026featcal}.
Training-based fusion adjusts the result through optimization, either by learning fusion coefficients, as in AdaMerging (Ada), or by training representation calibration modules, as in Surgery~\citep{yang2024adamerging,yang2024surgery}.
Distillation directly optimizes a student model: Seq-KD learns from teacher-generated responses, GKD provides teacher supervision on student-generated sequences, and MOPD uses on-policy distillation to integrate multiple domain teachers~\citep{kim2016seqkd,agarwal2024gkd,ma2026mopd}.

\paragraph{Model Fusion Evaluations.}
Existing model fusion evaluations differ in how they compare methods and select experts.
FusionBench supports comparisons across models and tasks through unified algorithm implementations, while Merging at Scale uses controlled experiments to study the effects of model size, base-model quality, and expert count~\citep{tang2024fusionbench,yadav2025matters}.
Expert sources further distinguish these settings: Model-GLUE studies the selection and combination of community models, whereas MergeBench trains domain experts under a common protocol to compare multi-task performance, pretrained knowledge retention, and runtime on consistent expert pools~\citep{zhao2024modelglue,he2025mergebench}.
Task coverage has also expanded from conventional NLP to math and code, but these settings mainly use task- or instruction-tuned experts, leaving capability retention after fusing long-CoT experts insufficiently studied~\citep{tang2024fusionbench,zhao2024modelglue,he2025mergebench}.
Tool use also receives limited coverage in existing model fusion evaluations, leaving a need for systematic comparisons of capability retention when independently trained long-CoT and tool use experts are fused.
\textsc{FusionEval} trains domain experts independently from a common base and compares fusion methods under consistent data-access conditions and evaluation tasks, reporting mean and worst-benchmark capability retention alongside computational cost.

\section{FusionEval}
\label{sec:benchmark}

\textsc{FusionEval} evaluates how well fused models retain domain-expert capabilities and how much computation fusion requires.
We select tasks in reasoning, instruction following, and tool use (Section~\ref{sec:task-construction}), and train domain experts independently from open-source base models (Section~\ref{sec:model-construction}).
Fusion methods use the same expert pool and are evaluated on eight public benchmarks under a common protocol (Section~\ref{sec:evaluation}).
We report mean and worst-benchmark normalized performance, runtime, and peak GPU memory (Section~\ref{sec:metrics}).

\subsection{Task Scope}
\label{sec:task-construction}

\textsc{FusionEval} covers five task categories: math, code, science question answering, instruction following, and tool use.
Math, code, and science question answering assess the reasoning capabilities of long-CoT experts; instruction following and tool use test compliance with explicit constraints and correct calls to external tools, respectively.
We examine whether fused models that retain math, code, and science reasoning capabilities can also follow instructions and call tools correctly, and which capabilities are more vulnerable to degradation during fusion.

\subsection{Expert Pool Construction}
\label{sec:model-construction}

\paragraph{Base models.}
We select advanced open-source models, Qwen3-4B-Base, Qwen3-8B-Base, and Llama-3.2-3B, as base models and construct an expert pool for each~\citep{yang2025qwen3,meta2024llama32}.
Each pool contains five experts trained independently by domain-specific SFT from the same base model, with the same architecture and tokenizer.
Fixing the data and training settings for each domain allows controlled expert construction and comparable fusion results.

\paragraph{Training data.}
\label{sec:data-construction}
Table~\ref{tab:training-data} summarizes the training datasets for each domain and the sample counts after preprocessing.
We apply basic data cleaning, including format standardization, deduplication, and filtering of overlong samples ($>32\mathrm{k}$ tokens) and abnormally repetitive content.
Before expert training, we remove training samples that exactly match evaluation samples in either raw or normalized text; Appendix Table~\ref{tab:data-decontamination} reports the counts.
Each domain expert is trained separately by SFT on its corresponding domain data; detailed training settings are provided in Appendix Table~\ref{tab:expert-sft-settings}.
We use prompts from the datasets listed in Appendix Table~\ref{tab:mopd-data} for distillation.
In MOPD, the student generates responses and the corresponding domain expert provides token-level supervision~\citep{ma2026mopd}; for Seq-KD, we train the student on responses generated in advance by domain experts~\citep{kim2016seqkd}.

\begin{table}[!t]
  \caption{Datasets for SFT of domain experts. Sample counts are reported after preprocessing.}
  \label{tab:training-data}
  \FusionEvalTableStyle
  \begin{tabularx}{\linewidth}{@{}l>{\raggedright\arraybackslash}Xr@{}}
    \toprule
    \textbf{Domain} & \textbf{Training datasets} & \textbf{\# Samples} \\
    \midrule
    \rowcolor{black!5}
    Math & \href{https://huggingface.co/datasets/OpenDataArena/ODA-Math-460k}{ODA-Math-460k}; \href{https://huggingface.co/datasets/Alibaba-Apsara/Superior-Reasoning-SFT-gpt-oss-120b}{Superior-Reasoning-SFT} (Math) & $334.1$K \\
    Code & \href{https://huggingface.co/datasets/nvidia/Nemotron-SFT-Competitive-Programming-v2}{Nemotron-Competitive-Programming-v2} & $300$K \\
    \rowcolor{black!5}
    Science & \href{https://huggingface.co/datasets/nvidia/OpenScienceReasoning-2}{OpenScienceReasoning-2} & $249.9$K \\
    IF & \href{https://huggingface.co/datasets/nvidia/Nemotron-SFT-Instruction-Following-Chat-v3}{Nemotron-Instruction-Following-Chat-v3} & $179.8$K \\
    \rowcolor{black!5}
    Agent & \href{https://huggingface.co/datasets/nvidia/Nemotron-Post-Training-Dataset-v1}{Nemotron-Post-Training-Dataset-v1}; \href{https://huggingface.co/datasets/Bingguang/HardGen}{HardGen}; \href{https://huggingface.co/datasets/Nanbeige/ToolMind}{ToolMind}; \href{https://huggingface.co/datasets/nvidia/Nemotron-Agentic-v1}{Nemotron-Agentic-v1} & $198.6$K \\
    \bottomrule
  \end{tabularx}
\end{table}
\FloatBarrier

\subsection{Evaluation Protocol}
\label{sec:evaluation}

We compare training-free and training-based fusion methods, including weight combination, data-assisted parameter fusion, and multi-teacher distillation. We also evaluate FeatCal and Surgery for post-fusion calibration. Table~\ref{tab:merging-methods} lists how each method computes the fused model and what data it requires; notation is defined in Appendix~\ref{app:notation}.

\begin{table}[htbp]
  \caption{Core operations and data requirements of evaluated fusion methods. Data access refers to model construction, excluding validation; inputs are unlabeled unless specified. RegMean is shown in its linear-layer form.}
  \label{tab:merging-methods}
  \FusionEvalTableStyle
  \begin{tabularx}{\linewidth}{@{}>{\raggedright\arraybackslash}p{0.13\linewidth} >{\raggedright\arraybackslash}X l@{}}
    \toprule
    \textbf{Method} & \textbf{Core operation} & \textbf{Data / signals} \\
    \midrule
    \multicolumn{3}{@{}l}{\textbf{Training-free}} \\
    Soup & $\theta^*=M^{-1}\sum_i\theta_i$ & Weights \\
    TA & $\theta^*=\theta_0+\lambda\sum_i\tau_i$ & Weights \\
    TIES & $\theta^*=\theta_0+\lambda\operatorname{TIES}_{\rho}(\{\tau_i\})$; trim/sign/mean & Weights \\
    DARE & $\theta^*=\theta_0+\lambda\sum_i\frac{(1-z_i)\odot\tau_i}{1-p}$; drop/rescale & Weights \\
    DELLA & Magnitude dropout, rescaling, sign-aligned mean & Weights \\
    L\&S & $\theta_j^*=\theta_{0j}+\dfrac{\sum_i m_{ij}\tau_{ij}}{\max(1,\sum_i m_{ij})}$ & Weights \\
    Fisher & $\theta^*=\dfrac{\sum_i F_i\odot\theta_i}{\sum_i F_i+\epsilon}$ & Inputs; Fisher estimates \\
    RegMean & $W^{*}=(\sum_i\widetilde{G}_i)^{-1}\sum_i\widetilde{G}_iW_i$ & Inputs; Gram matrices \\
    FeatCal & Forward-order regularized closed-form weight updates & Inputs; expert features \\
    \midrule
    \multicolumn{3}{@{}l}{\textbf{Training-based}} \\
    Ada & $(\theta^*)^\ell=\theta_0^\ell+\sum_i\lambda_i^\ell\tau_i^\ell$; entropy minimization & Unlabeled inputs \\
    Surgery & Train task-specific alignment modules & Inputs; expert features \\
    Seq-KD & Sequence imitation of teacher responses & Prompts; responses \\
    MOPD & Domain-teacher distillation on student rollouts & Prompts; token probabilities \\
    \bottomrule
  \end{tabularx}
\end{table}

We evaluate domain experts and fused models on eight public benchmarks across five domains; Table~\ref{tab:target-tasks} lists sample counts and generations per question.
Models from the same base use the same prompts, decoding parameters, and scoring rules on each benchmark.
Multi-turn tool tasks include issuing calls, receiving tool results, and continuing the interaction.
We select hyperparameters and checkpoints through quick evaluation on five benchmarks, then evaluate the selected models on all eight benchmarks.
Appendix~\ref{app:evaluation-protocols} details input templates, response extraction, and full settings.

\begin{table}[!t]
  \caption{Evaluation benchmarks by domain. $n$ denotes generations per question; slash-separated values follow the benchmark order.}
  \label{tab:target-tasks}
  \FusionEvalTableStyle
  \setlength{\tabcolsep}{4pt}
  \begin{tabular*}{\linewidth}{@{\extracolsep{\fill}}llll@{}}
    \toprule
    \textbf{Domain} & \textbf{Benchmarks} & \textbf{\# Samples} & \textbf{$n$} \\
    \midrule
    Math & AIME24 / AIME25~\citep{aime2025problems} & $30\,/\,30$ & $16\,/\,16$ \\
    Code & LCBv5 / LCBv6~\citep{jain2025livecodebench} & $279\,/\,131$ & $2\,/\,4$ \\
    Science & GPQA-D~\citep{rein2024gpqa} & $198$ & $4$ \\
    Instruction-following & IFEval / IFBench~\citep{zhou2023instruction,pyatkin2025generalizing} & $541\,/\,300$ & $2\,/\,2$ \\
    Agent & BFCLv3~\citep{patil2025bfcl} & $4{,}441$ & $1$ \\
    \bottomrule
  \end{tabular*}
\end{table}

\subsection{Metrics}
\label{sec:metrics}

\paragraph{Benchmark scores.}
AIME24, AIME25, and GPQA-D use answer accuracy; LCBv5 and LCBv6 use pass@1 based on program tests~\citep{jain2025livecodebench,rein2024gpqa}; IFEval and IFBench use prompt-level strict accuracy~\citep{zhou2023instruction,pyatkin2025generalizing}.
Repeated responses are scored individually and averaged, with generations per question listed in Table~\ref{tab:target-tasks}.
BFCLv3 uses micro-averaged success.
Raw Avg. is the equal-weight mean of the eight benchmark scores on the percentage scale.

\paragraph{Expert-normalized performance.}
We divide the score $s_b(\theta)$ of model $\theta$ on benchmark $b$ by the corresponding domain-expert score $s_b^{\mathrm{exp}}>0$ from the same pool, and compute the following over $K=8$ benchmarks~\citep{ilharco2023taskarithmetic,he2025mergebench}:
\begin{equation}
  \label{eq:normalized-performance}
  \mathrm{Avg.\,Norm.}
    =\frac{1}{K}\sum_{b=1}^{K}\frac{s_b(\theta)}{s_b^{\mathrm{exp}}},
  \qquad
  \mathrm{Min.\,Norm.}
    =\min_{1\le b\le K}\frac{s_b(\theta)}{s_b^{\mathrm{exp}}}.
\end{equation}
Avg. Norm. measures mean performance relative to the experts, while Min. Norm. measures normalized performance on the worst benchmark to identify capability losses that the mean may hide.
A normalized score of $1$ matches the corresponding expert.

\paragraph{Computational cost.}
Following MergeBench~\citep{he2025mergebench}, we measure Algorithm runtime and Peak GPU memory during fusion to assess computational cost.
Appendix~\ref{app:data-access-efficiency} gives the measurement settings; Appendix~\ref{app:hyperparameter-tuning} reports search counts and cost estimates.

\section{Experiments}
\label{sec:fusioneval-evaluation}

We first compare multi-domain performance on the Qwen3-4B and Qwen3-8B expert pools, focusing on mean performance and capability retention on the worst benchmark (Section~\ref{sec:main-fusion-levels}). We then examine Continual Fusion (Section~\ref{sec:continual-model-merging}), and analyze fusion costs, parameter update patterns, and the effects of post-fusion calibration (Section~\ref{sec:discussion-analysis}).

\subsection{Multi-Domain Fusion Performance}
\label{sec:main-fusion-levels}

We compare multi-domain fusion performance on the Qwen3-4B and Qwen3-8B expert pools (Tables~\ref{tab:qwen3-4b-fusion-results} and~\ref{tab:parameter-level-8b-results}). We also evaluate the Llama-3.2-3B expert pool to examine whether the findings hold in another model family; settings and results are provided in Appendix~\ref{app:llama-evaluation}. Data-assisted fusion uses 128 calibration examples per domain. We randomly sample 70\% of each domain's training data and combine them for SFT of the base model, producing Student as the distillation initialization and a comparison baseline. Distillation data and training configurations are provided in Appendices~\ref{app:distillation-data} and~\ref{app:selected-fusion-configurations}. Non-distillation methods are ordered by Min.\ Norm., with distillation methods listed separately. Appendices~\ref{app:hyperparameter-tuning} and~\ref{app:selected-fusion-configurations} give search settings and method configurations; Appendix~\ref{app:expert-pool-results} reports cross-domain expert results, and Appendix~\ref{app:additional-fusion-results} gives full LOT and post-fusion calibration results.

\begin{table}[ht]
  \caption{Model fusion performance on Qwen3-4B. Benchmark scores and Raw Avg. are percentages; Avg. Norm. and Min. Norm. are relative to the corresponding domain experts. Experts lists the per-benchmark reference scores, and Student is the mixed-domain SFT baseline. Bold marks the best fusion result in each column.}
  \label{tab:qwen3-4b-fusion-results}
  \FusionEvalTableStyle
  \setlength{\tabcolsep}{2.4pt}
  \begin{tabular}{@{}l*{11}{c}@{}}
    \toprule
    & \multicolumn{2}{c}{\textbf{Math}} & \multicolumn{2}{c}{\textbf{Code}} & \textbf{Science} & \multicolumn{2}{c}{\textbf{IF}} & \textbf{Tool use} & \textbf{Raw} & \multicolumn{2}{c}{\textbf{Norm.}} \\
    \cmidrule(lr){2-3}\cmidrule(lr){4-5}\cmidrule(lr){6-6}\cmidrule(lr){7-8}\cmidrule(lr){9-9}\cmidrule(l){11-12}
 % Fusion rows are ordered by Min. Norm.; incomplete summaries follow.
\textbf{Method} & \textbf{AIME24} & \textbf{AIME25} & \textbf{LCBv5} & \textbf{LCBv6} & \textbf{GPQA-D} & \textbf{IFEval} & \textbf{IFBench} & \textbf{BFCLv3} & \textbf{Avg.} & \textbf{Avg.} & \textbf{Min.} \\
    \midrule

    Experts & 73.4 & 67.4 & 44.4 & 41.6 & 51.9 & 86.0 & 38.6 & 66.0 & 58.7 & 1.000 & 1.000 \\
    \midrule
    TIES & \textbf{69.0} & 64.4 & 41.6 & 35.9 & 36.9 & 65.3 & 36.2 & 30.0 & 47.4 & 0.819 & 0.454 \\
    RegMean & 43.5 & 40.2 & 31.5 & 29.4 & 47.1 & 68.9 & 28.7 & 30.4 & 40.0 & 0.690 & 0.460 \\
DARE & 66.9 & 62.5 & 47.5 & \textbf{42.6} & 49.0 & 70.9 & 37.0 & 30.6 & 50.9 & 0.890 & 0.464 \\
    Ada & 67.5 & 58.8 & 42.3 & 38.7 & 49.6 & 66.6 & 35.0 & 31.0 & 48.7 & 0.848 & 0.470 \\
DELLA & 68.5 & 64.8 & 47.0 & 40.8 & 49.5 & 64.4 & 36.0 & 31.1 & 50.3 & 0.880 & 0.471 \\
TA & 67.5 & \textbf{67.1} & 48.2 & 42.2 & \textbf{51.4} & 69.4 & 37.8 & 31.1 & 51.8 & \textbf{0.908} & 0.472 \\
L\&S & 67.1 & 62.5 & \textbf{49.3} & 39.5 & 44.6 & 70.3 & \textbf{42.3} & 31.5 & 50.9 & 0.894 & 0.478 \\
Soup & 45.4 & 35.6 & 27.8 & 27.5 & 51.3 & 62.9 & 24.8 & 33.8 & 38.6 & 0.663 & 0.512 \\
    Fisher & 39.8 & 38.1 & 38.2 & 31.9 & 47.1 & 63.9 & 29.6 & 45.1 & 41.7 & 0.729 & 0.542 \\
    \midrule
    \rowcolor{black!5}
Student & 49.1 & 47.9 & 33.5 & 31.5 & 30.9 & 64.8 & 27.8 & 63.7 & 43.7 & 0.741 & 0.596 \\
    Seq-KD & 47.9 & 43.8 & 36.4 & 31.9 & 34.5 & 67.7 & 37.5 & 63.7 & 45.4 & 0.784 & 0.649 \\
MOPD & 59.6 & 53.3 & 42.8 & 38.9 & 36.7 & \textbf{77.5} & 41.4 & \textbf{66.1} & \textbf{52.1} & 0.899 & \textbf{0.708} \\
    \bottomrule
  \end{tabular}
\end{table}
\FloatBarrier

\begin{table}[ht]
  \caption{Model fusion performance on Qwen3-8B. Benchmark scores and Raw Avg. are percentages; Avg. Norm. and Min. Norm. are relative to the corresponding domain experts. Experts lists the per-benchmark reference scores, and Student is the mixed-domain SFT baseline. Bold marks the best fusion result in each column.}
  \label{tab:parameter-level-8b-results}
  \FusionEvalTableStyle
  \setlength{\tabcolsep}{2.4pt}
  \begin{tabular}{@{}l*{11}{c}@{}}
    \toprule
    & \multicolumn{2}{c}{\textbf{Math}} & \multicolumn{2}{c}{\textbf{Code}} & \textbf{Science} & \multicolumn{2}{c}{\textbf{IF}} & \textbf{Tool use} & \textbf{Raw} & \multicolumn{2}{c}{\textbf{Norm.}} \\
    \cmidrule(lr){2-3}\cmidrule(lr){4-5}\cmidrule(lr){6-6}\cmidrule(lr){7-8}\cmidrule(lr){9-9}\cmidrule(l){11-12}
 \textbf{Method} & \textbf{AIME24} & \textbf{AIME25} & \textbf{LCBv5} & \textbf{LCBv6} & \textbf{GPQA-D} & \textbf{IFEval} & \textbf{IFBench} & \textbf{BFCLv3} & \textbf{Avg.} & \textbf{Avg.} & \textbf{Min.} \\
    \midrule

    Experts & 74.6 & 67.9 & 47.3 & 43.5 & 52.9 & 84.6 & 42.0 & 67.5 & 60.0 & 1.000 & 1.000 \\
    \midrule
    Fisher & 60.4 & 41.0 & 30.6 & 28.8 & 45.2 & 64.0 & 26.8 & 29.8 & 40.8 & 0.677 & 0.442 \\
    Ada & 69.0 & 65.8 & 55.9 & 48.5 & 48.2 & 73.9 & 41.4 & 31.3 & 54.3 & 0.928 & 0.463 \\
 TA & 69.8 & 68.1 & 56.6 & \textbf{52.7} & 55.4 & 74.5 & 39.8 & 31.5 & 56.1 & 0.961 & 0.466 \\
    DARE & 71.9 & 67.1 & 57.5 & 48.7 & 55.6 & 74.7 & 35.2 & 31.8 & 55.3 & 0.941 & 0.471 \\
    RegMean & 52.5 & 42.1 & 37.1 & 35.7 & 54.0 & 72.3 & 29.0 & 31.8 & 44.3 & 0.746 & 0.471 \\
 Soup & 53.1 & 39.8 & 41.2 & 36.1 & 57.2 & 75.0 & 31.8 & 34.1 & 46.0 & 0.778 & 0.505 \\
    TIES & 69.0 & 63.5 & 58.4 & 44.5 & 52.3 & 73.7 & 43.2 & 34.8 & 54.9 & 0.940 & 0.515 \\
 DELLA & \textbf{75.6} & 69.4 & \textbf{59.7} & 51.3 & \textbf{58.2} & 71.9 & 39.2 & 37.8 & \textbf{57.9} & \textbf{0.990} & 0.561 \\
 L\&S & 72.7 & \textbf{72.1} & 55.4 & 50.8 & 51.6 & 75.1 & 42.0 & 39.1 & 57.4 & 0.977 & 0.580 \\
    \midrule
    \rowcolor{black!5}
    Student & 52.8 & 51.5 & 39.2 & 36.2 & 31.2 & 67.4 & 30.8 & 63.1 & 46.5 & 0.773 & 0.590 \\
    Seq-KD & 45.8 & 55.8 & 37.1 & 34.9 & 44.4 & 68.5 & 40.7 & 65.0 & 49.0 & 0.826 & 0.615 \\
    MOPD & 67.9 & 59.8 & 45.9 & 42.6 & 45.5 & \textbf{80.3} & \textbf{46.1} & \textbf{66.9} & 56.9 & 0.955 & \textbf{0.859} \\
    \bottomrule
  \end{tabular}
\end{table}
\FloatBarrier

\label{sec:capability-tradeoffs}
\paragraph{Training-free methods fall short in retaining tool use.}
On 4B, BFCLv3 is the weakest normalized benchmark for seven of the eight training-free methods. TA scores $31.1$ on BFCLv3, close to the $29.7$--$29.8$ range of experts from other domains and well below the tool use expert at $66.0$ (Appendix Table~\ref{tab:pilot-expert-results}). On 8B, BFCLv3 is the weakest normalized benchmark for all eight methods; even L\&S, the best of these methods on tool use, scores only $39.1$, compared with $67.5$ for the corresponding expert.

\paragraph{MOPD achieves more balanced performance across domains.}
MOPD has $0.899$ and $0.955$ Avg.\ Norm. on 4B and 8B, below the best score in each pool, but the highest Min.\ Norm. among evaluated methods at $0.708$ and $0.859$. Student already has strong performance in tool use, which MOPD retains while scoring higher in reasoning and instruction following. In contrast, Seq-KD scores higher than Student on average but lower on some math or code benchmarks, and trails MOPD on all eight benchmarks in both pools.

\paragraph{Larger model scales do not guarantee better capability retention.}
In both the 4B and 8B experiments, several methods retain reasoning performance well, but tool use remains the least retained capability for most non-distillation methods. TA's Avg.\ Norm. rises from $0.908$ on 4B to $0.961$ on 8B, but its Min.\ Norm. falls from $0.472$ to $0.466$. DELLA and L\&S improve on both metrics, yet their 8B scores on tool use remain at only about $56\%$ and $58\%$ of the corresponding expert score. Although some methods achieve higher normalized scores on 8B, the difference in retention between reasoning and tool use remains.

\subsection{Continual Fusion}
\label{sec:continual-model-merging}
Expert models often become available sequentially in practice, a setting captured by Continual Fusion~\citep{tang2025opcm,yang2025dop}.
On Qwen3-4B, we evaluate TA, Continual Soup, and Continual DELLA, starting from the instruction-following expert and successively fusing the math, science, code, and tool use experts to examine capability acquisition and retention.
Each step uses the current fused model and the new expert, without access to earlier experts' individual weights.
The common base model, calibration data from previously introduced domains, and retained statistics remain available.
Table~\ref{tab:continual-results} reports stage-wise results, with joint fusion of all five experts as a reference for Soup and DELLA.
Appendix~\ref{app:continual-results} details differences in initialization, effective coefficients, and statistical configurations between continual and joint fusion.

\begin{table}[htbp]
  \caption{Continual Fusion on Qwen3-4B, starting from the IF expert (Avg.\ Norm.\ $0.497$, Min.\ Norm.\ $0.227$) and adding one domain expert per stage. Avg. and Min. are expert-normalized scores across eight benchmarks.}
  \label{tab:continual-results}
  \FusionEvalTableStyle
  \setlength{\tabcolsep}{1.6pt}
  \begin{tabular*}{\linewidth}{@{\extracolsep{\fill}}l*{10}{c}@{}}
    \toprule
    \textbf{Method} & \multicolumn{2}{c}{\textbf{1: +Math}} & \multicolumn{2}{c}{\textbf{2: +Science}} & \multicolumn{2}{c}{\textbf{3: +Code}} & \multicolumn{2}{c}{\textbf{4: +Agent}} & \multicolumn{2}{c}{\textbf{Joint fusion}} \\
    \cmidrule(lr){2-3}\cmidrule(lr){4-5}\cmidrule(lr){6-7}\cmidrule(lr){8-9}\cmidrule(l){10-11}
    & \textbf{Avg.} & \textbf{Min.} & \textbf{Avg.} & \textbf{Min.} & \textbf{Avg.} & \textbf{Min.} & \textbf{Avg.} & \textbf{Min.} & \textbf{Avg.} & \textbf{Min.} \\
    \midrule
    TA & 0.463 & 0.303 & 0.616 & 0.402 & 0.814 & 0.450 & 0.884 & 0.468 & - & - \\
    \midrule
    Continual Soup & 0.624 & 0.451 & 0.668 & 0.451 & 0.744 & 0.451 & 0.583 & 0.352 & 0.663\,($-12.1\%$) & 0.512\,($-31.2\%$) \\
    Continual TIES & 0.298 & 0.056 & 0.499 & 0.178 & 0.396 & 0.161 & 0.642 & 0.495 & 0.819\,($-21.6\%$) & 0.454\,($+9.0\%$) \\
    Continual DELLA & 0.705 & 0.451 & 0.731 & 0.449 & 0.858 & 0.452 & 0.776 & 0.460 & 0.880\,($-11.8\%$) & 0.471\,($-2.3\%$) \\
    \bottomrule
  \end{tabular*}
\end{table}
\FloatBarrier

\paragraph{Continual Fusion still falls short of joint fusion.}
Under the current order and configurations, Continual Soup and DELLA finish $12.1\%$ and $11.8\%$ below their respective joint fusion references in normalized average performance.
Adding the science expert improves math performance for both methods, and adding the code expert further improves it, showing that new experts can strengthen existing capabilities.
However, adding the tool use expert yields little improvement in tool use while degrading math, code, and instruction following, as shown in Appendix Table~\ref{tab:continual-final-scores}.
Their normalized average scores consequently decrease from $0.744$ to $0.583$ and from $0.858$ to $0.776$, respectively.
DELLA also retains a minimum normalized score closer to its joint fusion reference, with a relative decline of $2.3\%$ compared with $31.2\%$ for Soup.
Overall, improvements in individual capabilities do not translate into average performance matching joint fusion, highlighting the challenge of acquiring new capabilities while retaining existing ones.

\subsection{Analysis}
\label{sec:discussion-analysis}

\begin{figure}[!htbp]
  \centering
  \includegraphics[width=\linewidth]{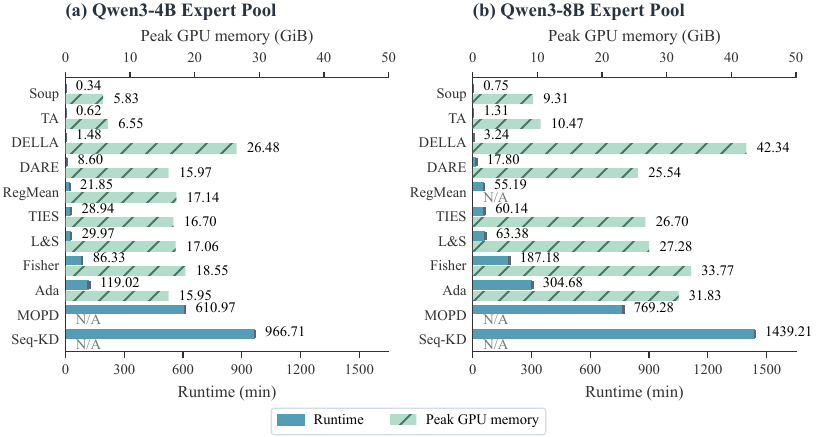}
  \caption{Runtime and peak GPU memory on Qwen3-4B and Qwen3-8B. Blue bars show mean runtime (bottom axis); hatched green bars show peak memory (top axis). Error bars indicate runtime ranges; N/A denotes unrecorded memory. Resource and timing details appear in Appendix~\ref{app:data-access-efficiency}.}
  \label{fig:efficiency-retest}
\end{figure}

\paragraph{Computational cost.}
\label{sec:efficiency}
Figure~\ref{fig:efficiency-retest} reports runtime and peak GPU memory for the main experiments. On Qwen3-4B, TA takes $0.62$ minutes and achieves higher Avg.\ Norm. than the slower DELLA and L\&S, but retains tool use poorly. DELLA runs faster than TIES but requires more peak GPU memory, at $26.48$ versus $16.70$ GiB. Among the distillation methods, MOPD achieves higher Min.\ Norm., with training taking about $10.18$ and $12.82$ hours on 4B and 8B. Seq-KD takes about $16.11$ and $23.99$ hours, including offline teacher response generation and student training. Appendix~\ref{app:data-access-efficiency} gives the resource configurations and timing scope for each method.

\FloatBarrier
\vspace{-6pt}
\setlength{\intextsep}{1pt}
\paragraph{Examining Tool Use Degradation.}
\label{sec:mechanistic-analysis}
\begin{wrapfigure}{r}{0.49\textwidth}
  \setlength{\abovecaptionskip}{1pt}
  \centering
  \includegraphics[width=\linewidth]{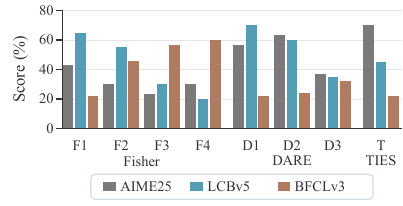}
  \caption{AIME25, LCBv5, and BFCLv3 \mbox{screening} scores on Qwen3-4B.}
  \label{fig:tool-tradeoff}
\end{wrapfigure}
\leavevmode\newline
We examine whether update norm, Agent cosine similarity, and Agent-weighted sign conflict consistently track BFCLv3 screening performance (Figure~\ref{fig:tool-degradation}).
Within Fisher, higher BFCLv3 scores coincide with smaller update norms and higher Agent cosine similarity.
Across methods, however, TIES has the largest norm but performs similarly to F1, while F2 outperforms D1 despite lower Agent cosine similarity.
TIES also has lower weighted sign conflict than D1/D2 without higher BFCLv3 scores.
These comparisons caution against selecting configurations based on a single update statistic.
Figure~\ref{fig:tool-tradeoff} further shows capability tradeoffs: F2--F4 outperform F1 on BFCLv3 but score lower on AIME25 and LCBv5, with the same pattern for D3 relative to D1/D2.
These changes are not uniform across reasoning domains: from F3 to F4, BFCLv3 and AIME25 both improve, whereas LCBv5 declines further.
Higher tool use scores therefore do not by themselves establish better joint capability retention.
Configuration selection should assess tool use and individual reasoning domains jointly rather than prioritize BFCLv3 or a single update statistic alone.
Appendix~\ref{app:tool-degradation} provides configurations, results, and a case study.

\setlength{\intextsep}{12pt plus 2pt minus 2pt}

\begingroup
\setlength{\intextsep}{2pt}
\setlength{\textfloatsep}{4pt plus 1pt minus 2pt}
\setlength{\abovecaptionskip}{1pt}
\setlength{\belowcaptionskip}{0pt}
\begin{figure}[H]
  \centering
  \includegraphics[width=\linewidth]{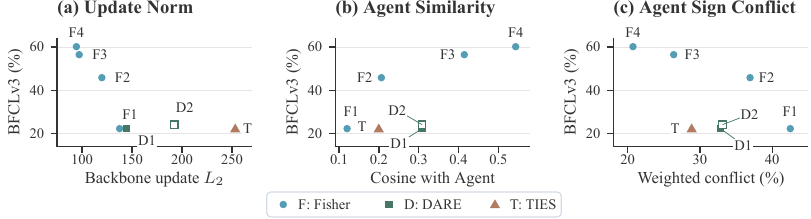}
  \caption{BFCLv3 screening performance versus update norm, Agent cosine similarity, and Agent-weighted sign conflict on Qwen3-4B. F1--F4: Fisher; D1/D2: DARE (drop rate $0.2$, scales $0.3/0.4$, respectively); T: TIES. See Appendix~\ref{app:tool-degradation} for definitions and additional configurations.}  \label{fig:tool-degradation}
\end{figure}
\endgroup

\FloatBarrier

\begin{wraptable}{r}{0.49\textwidth}
  \centering
  \caption{Effect of calibrating TA with FeatCal or Surgery. Values are changes in normalized average and worst-benchmark scores relative to TA.}
  \label{tab:post-fusion-calibration}
  \FusionEvalTableStyle
  \begin{tabular*}{\linewidth}{@{\extracolsep{\fill}}llrr@{}}
    \toprule
    & & \multicolumn{2}{c}{\textbf{Norm.}} \\
    \cmidrule(l){3-4}
    \textbf{Model} & \textbf{Calibration} & \textbf{Avg.} & \textbf{Min.} \\
    \midrule
    \multirow{2}{*}{Qwen3-4B} & FeatCal & $-0.039$ & $+0.011$ \\
     & Surgery & $-0.023$ & $+0.003$ \\
    \midrule
    \multirow{2}{*}{Qwen3-8B} & FeatCal & $-0.002$ & $-0.001$ \\
     & Surgery & $-0.068$ & $+0.042$ \\
    \bottomrule
  \end{tabular*}
\end{wraptable}

\paragraph{Post-Fusion Calibration.}
\label{sec:post-fusion-calibration}
This analysis evaluates the effects of post-fusion calibration on average and worst-benchmark performance. Table~\ref{tab:post-fusion-calibration} compares FeatCal and Surgery against TA. On 4B, Avg.\ Norm. falls by 0.039 and 0.023, respectively, while Min.\ Norm. rises by only 0.011 and 0.003. On 8B, FeatCal changes neither metric appreciably; Surgery raises Min.\ Norm. by 0.042 but lowers Avg.\ Norm. by 0.068. Gains on the weakest benchmark thus coexist with losses elsewhere, and neither calibration improves both metrics under these configurations. Full results appear in Appendix~\ref{app:additional-fusion-results}.

\FloatBarrier

\subsection{Future Directions}
\label{sec:future-directions}

\paragraph{Preserving reasoning and tool use in training-free fusion.}
Our results show that several training-free methods retain much of the math and code performance but substantially degrade tool use.
Update magnitude, similarity, and sign conflict do not fully explain these changes individually.
Future work should test whether preserving updates from the tool use expert in selected layers or parameter groups can reduce these losses without sacrificing reasoning.
Selection could be guided by capability retention on separate validation data rather than weight statistics alone.
The goal is joint retention without additional training, not higher average scores with degraded tool use.

\paragraph{MOPD is effective but costly.}
MOPD achieves a more balanced capability profile and retains tool use, but requires additional data preparation, inference, and student training compared with training-free fusion. 
Since the student already performs well on tool use before distillation, future work could select teachers and prompts for weaker domains while checking that this existing strength is preserved. 
Another direction is to combine parameter fusion with selective distillation to recover degraded capabilities using fewer teacher queries and training steps. 
These approaches should be evaluated by total preparation, inference, and training costs, alongside capability retention and gains over the initial student.

\paragraph{Continual model fusion remains challenging.}
All three continual methods finish below joint fusion in average performance, and adding the tool use expert yields little improvement in tool use for Continual Soup and DELLA. 
Future work should test whether preserving updates from the tool use expert in selected layers or parameter groups can reduce these losses without sacrificing reasoning, with global reweighting of that expert as a baseline.

\FloatBarrier

\section{Conclusion}
\label{sec:conclusion}

We introduce \textsc{FusionEval} to address the limited coverage of long-chain reasoning and tool use in existing model fusion evaluations. \textsc{FusionEval} provides five-domain expert pools built from Qwen3 and Llama models and compares training-free and training-based methods under shared data-access rules and eight public benchmarks, measuring capability retention, runtime, and peak GPU memory. Several methods retain strong reasoning performance but incur substantial losses in tool use, showing that high average scores do not establish joint retention across domains. MOPD retains tool use while achieving more balanced performance, but requires additional data preparation, inference, and training. Continual Fusion can strengthen some existing capabilities as new experts arrive, but leads to an overall loss of model capabilities relative to joint fusion in our experiments. We will release the data, expert pools, model fusion scripts, and related configurations to support comparisons under the same protocol. We hope \textsc{FusionEval} supports the development of parameter- and data-efficient fusion methods that retain expert capabilities more fully at lower computational cost, with attention to tool use, which is vulnerable to degradation during fusion.

\bibliography{references}
\bibliographystyle{iclr2027_conference}

\appendix
\raggedbottom
\makeatletter
\setlength{\@fptop}{0pt}
\setlength{\@fpsep}{12pt}
\setlength{\@fpbot}{0pt plus 1fil}
\makeatother
\section{Expert Pool Results}
\label{app:expert-pool-results}

\begin{table}[H]
  \caption{Cross-domain expert scores (\%) on Qwen3-4B. Bold marks column maxima.}
  \label{tab:pilot-expert-results}
  \FusionEvalTableStyle
  \begin{tabular*}{\linewidth}{@{\extracolsep{\fill}}l*{8}{c}@{}}
    \toprule
    & \multicolumn{2}{c}{\textbf{Math}} & \multicolumn{2}{c}{\textbf{Code}} & \textbf{Science} & \multicolumn{2}{c}{\textbf{IF}} & \textbf{Tool use} \\
    \cmidrule(lr){2-3}\cmidrule(lr){4-5}\cmidrule(lr){6-6}\cmidrule(lr){7-8}\cmidrule(l){9-9}
\textbf{Expert} & \textbf{AIME24} & \textbf{AIME25} & \textbf{LCBv5} & \textbf{LCBv6} & \textbf{GPQA-D} & \textbf{IFEval} & \textbf{IFBench} & \textbf{BFCLv3} \\
    \midrule
    Math & \textbf{73.4} & \textbf{67.4} & 21.7 & 22.0 & 39.7 & 48.3 & 21.0 & 29.7 \\
    Code & 46.5 & 45.4 & \textbf{44.4} & \textbf{41.6} & 31.8 & 58.7 & 27.3 & 29.7 \\
    Science & 47.5 & 45.2 & 15.6 & 15.8 & \textbf{51.9} & 44.6 & 20.8 & 29.8 \\
    IF & 16.7 & 17.1 & 14.2 & 14.1 & 20.2 & \textbf{86.0} & \textbf{38.6} & 29.8 \\
    Agent & 17.1 & 13.3 & 10.8 & 13.9 & 31.7 & 51.7 & 23.1 & \textbf{66.0} \\
    \bottomrule
  \end{tabular*}
\end{table}

\begin{table}[H]
  \caption{Cross-domain expert scores (\%) on Qwen3-8B. Bold marks column maxima.}
  \label{tab:qwen3-8b-expert-results}
  \FusionEvalTableStyle
  \begin{tabular*}{\linewidth}{@{\extracolsep{\fill}}l*{8}{c}@{}}
    \toprule
    & \multicolumn{2}{c}{\textbf{Math}} & \multicolumn{2}{c}{\textbf{Code}} & \textbf{Science} & \multicolumn{2}{c}{\textbf{IF}} & \textbf{Tool use} \\
    \cmidrule(lr){2-3}\cmidrule(lr){4-5}\cmidrule(lr){6-6}\cmidrule(lr){7-8}\cmidrule(l){9-9}
 \textbf{Expert} & \textbf{AIME24} & \textbf{AIME25} & \textbf{LCBv5} & \textbf{LCBv6} & \textbf{GPQA-D} & \textbf{IFEval} & \textbf{IFBench} & \textbf{BFCLv3} \\
    \midrule
    Math & \textbf{74.6} & \textbf{67.9} & 28.5 & 25.6 & 41.9 & 60.9 & 24.8 & 29.7 \\
    Code & 58.8 & 57.9 & \textbf{47.3} & \textbf{43.5} & 42.9 & 68.6 & 31.7 & 29.3 \\
    Science & 40.6 & 35.4 & 24.2 & 25.0 & \textbf{52.9} & 54.0 & 21.3 & 28.4 \\
    IF & 29.4 & 24.0 & 22.0 & 24.2 & 40.2 & \textbf{84.6} & \textbf{42.0} & 29.8 \\
    Agent & 22.3 & 20.6 & 14.9 & 15.7 & 46.7 & 62.0 & 23.0 & \textbf{67.5} \\
    \bottomrule
  \end{tabular*}
\end{table}

\subsection{Llama-3.2-3B Evaluation}
\label{app:llama-evaluation}

To examine whether the Qwen3 findings hold in another model family, we compare TA, TIES, DARE, DELLA, Consensus TA~\citep{wang2024localizing}, and TA+FeatCal on the Llama-3.2-3B expert pool. Table~\ref{tab:llama3-3b-fusion-results} reports the results, with method configurations in Appendix Table~\ref{tab:selected-config-llama}.

\begin{table}[H]
  \caption{Cross-domain expert scores (\%) on Llama-3.2-3B. Bold marks column maxima.}
  \label{tab:llama3-3b-expert-results}
  \FusionEvalTableStyle
  \begin{tabular*}{\linewidth}{@{\extracolsep{\fill}}l*{8}{c}@{}}
    \toprule
    & \multicolumn{2}{c}{\textbf{Math}} & \multicolumn{2}{c}{\textbf{Code}} & \textbf{Science} & \multicolumn{2}{c}{\textbf{IF}} & \textbf{Tool use} \\
    \cmidrule(lr){2-3}\cmidrule(lr){4-5}\cmidrule(lr){6-6}\cmidrule(lr){7-8}\cmidrule(l){9-9}
 \textbf{Expert} & \textbf{AIME24} & \textbf{AIME25} & \textbf{LCBv5} & \textbf{LCBv6} & \textbf{GPQA-D} & \textbf{IFEval} & \textbf{IFBench} & \textbf{BFCLv3} \\
    \midrule
    Math & \textbf{32.2} & \textbf{31.2} & 2.1 & 2.6 & 11.6 & 22.8 & 13.1 & 21.0 \\
    Code & 2.2 & 1.3 & \textbf{21.8} & \textbf{23.1} & 8.3 & 34.5 & 19.0 & 7.8 \\
    Science & 5.0 & 2.1 & 0.9 & 1.9 & \textbf{35.0} & 25.3 & 13.2 & 8.8 \\
    IF & 0.2 & 0.2 & 1.3 & 1.9 & 9.7 & \textbf{77.5} & \textbf{29.8} & 4.1 \\
    Agent & 0.0 & 0.0 & 0.9 & 1.3 & 23.9 & 38.4 & 13.4 & \textbf{62.6} \\
    \bottomrule
  \end{tabular*}
\end{table}

As in the Qwen3 parameter fusion experiments, tool use remains the least retained capability for most methods, with BFCLv3 giving the lowest normalized score for five of the six methods. TIES has the highest score on tool use among the six methods but the lowest scores on both code benchmarks, again showing a trade-off between reasoning and tool use. The difference is in reasoning retention. Some fused models outperform the corresponding experts on reasoning benchmarks in the Qwen3 experiments, whereas all six methods fall below the corresponding domain experts on all eight benchmarks on Llama, with substantial losses in math, code, and science reasoning as well. DELLA, for example, exceeds the experts on all five reasoning benchmarks on Qwen3-8B but falls below them on all five on Llama. Although it still has the highest Avg.\ Norm. among the six methods on Llama, the score is only $0.568$, with a Min.\ Norm. of $0.114$.

\begin{table}[H]
  \caption{Model fusion on Llama-3.2-3B. Benchmark scores and Raw Avg. are percentages; Norm. scores are relative to the domain experts. Bold marks the best fusion result per column.}
  \label{tab:llama3-3b-fusion-results}
  \FusionEvalTableStyle
  \setlength{\tabcolsep}{1.6pt}
  \begin{tabular}{@{}l*{11}{c}@{}}
    \toprule
    & \multicolumn{2}{c}{\textbf{Math}} & \multicolumn{2}{c}{\textbf{Code}} & \textbf{Science} & \multicolumn{2}{c}{\textbf{IF}} & \textbf{Tool use} & \textbf{Raw} & \multicolumn{2}{c}{\textbf{Norm.}} \\
    \cmidrule(lr){2-3}\cmidrule(lr){4-5}\cmidrule(lr){6-6}\cmidrule(lr){7-8}\cmidrule(lr){9-9}\cmidrule(l){11-12}
 \textbf{Method} & \textbf{AIME24} & \textbf{AIME25} & \textbf{LCBv5} & \textbf{LCBv6} & \textbf{GPQA-D} & \textbf{IFEval} & \textbf{IFBench} & \textbf{BFCLv3} & \textbf{Avg.} & \textbf{Avg.} & \textbf{Min.} \\
    \midrule

    Experts & 32.2 & 31.2 & 21.8 & 23.1 & 35.0 & 77.5 & 29.8 & 62.6 & 39.1 & 1.000 & 1.000 \\
    \midrule
    DELLA & \textbf{25.0} & 17.7 & 13.8 & 13.2 & \textbf{22.6} & 45.6 & 19.4 & 7.1 & 20.5 & \textbf{0.568} & 0.114\\
    TIES & 13.5 & 14.8 & 3.9 & 3.1 & 11.6 & 36.9 & 17.0 & \textbf{15.7} & 14.6 & 0.355 & 0.132\\
    TA & 20.0 & 15.8 & 10.8 & 10.9 & 16.9 & 50.6 & 20.2 & 8.9 & 19.3 & 0.506 & 0.142\\
    Consensus TA & 16.5 & 14.2 & 12.2 & 11.6 & 17.3 & \textbf{52.2} & 23.2 & 9.2 & 19.5 & 0.515 & 0.147\\
    DARE & 20.6 & 14.2 & \textbf{15.6} & \textbf{15.1} & 19.7 & 48.9 & 21.3 & 9.5 & \textbf{20.6} & 0.565 & 0.152\\
    TA+FeatCal & 21.9 & \textbf{22.9} & 8.6 & 9.2 & 16.0 & 46.6 & \textbf{23.4} & 14.9 & 20.4 & 0.536 & \textbf{0.238}\\
    \bottomrule
  \end{tabular}
\end{table}

\section{Implementation Details}
\label{app:implementation-details}

\subsection{Expert SFT Settings}
\label{app:expert-sft-settings}

\begin{table}[H]
  \caption{Target training settings for domain-expert SFT.}
  \label{tab:expert-sft-settings}
  \FusionEvalTableStyle
  \begin{tabular}{@{}ll@{}}
    \toprule
    \textbf{Setting} & \textbf{Value} \\
    \midrule
    Optimizer & Adam ($\beta_1=0.9$, $\beta_2=0.95$) \\
    Peak learning rate (target) & $5\times10^{-5}$ \\
    Global batch size (target) & $256$ \\
    Training epochs (target) & $5$ \\
    Learning-rate schedule & Warmup followed by cosine decay \\
    Loss & SFT loss averaged over supervised tokens \\
    Precision & BF16 \\
    \bottomrule
  \end{tabular}
\end{table}

\subsection{Notation}
\label{app:notation}

\begin{table}[H]
  \caption{Notation used in the main text.}
  \label{tab:notation}
  \FusionEvalTableStyle
  \begin{tabularx}{\linewidth}{@{}l >{\raggedright\arraybackslash}X@{}}
    \toprule
    \textbf{Symbol} & \textbf{Definition} \\
    \midrule
    $M,\ i$ & Number of experts and expert index ($M=5$). \\
    $K,\ b$ & Number of benchmarks and benchmark index ($K=8$). \\
    $n$ & Generations per question. \\
    $\theta_0,\ \theta_i$ & Base-model and expert parameters. \\
    $\theta,\ \theta^*$ & Evaluated model and fused-model parameters. \\
    $\tau_i$ & Expert task vector, $\theta_i-\theta_0$. \\
    $j,\ \ell$ & Parameter coordinate and layer indices. \\
    $s_b(\theta),\ s_b^{\mathrm{exp}}$ & Model and domain-expert scores on benchmark $b$. \\
    $\mathrm{Raw\,Avg.}$ & Mean benchmark score (\%). \\
    $\mathrm{Avg.\,Norm.},\ \mathrm{Min.\,Norm.}$ & Mean and minimum expert-normalized scores. \\
    $\lambda,\ \lambda_i^\ell$ & Global and per-expert, per-layer fusion coefficients. \\
    $\rho,\ \operatorname{TIES}_{\rho}$ & Retained fraction and TIES aggregation operator. \\
    $p,\ z_i$ & Drop probability and binary drop mask ($1$: dropped). \\
    $F_i,\ \epsilon$ & Fisher weights and stability constant. \\
    $m_{ij}$ & Localization mask ($1$: retained). \\
    $W_i,\ W^*$ & Expert and fused linear-layer weights. \\
    $G_i,\ \widetilde G_i$ & Layer-input Gram matrix and its regularized form. \\
    $\odot$ & Element-wise multiplication. \\
    \bottomrule
  \end{tabularx}
\end{table}

\subsection{Data Decontamination}
\label{app:data-decontamination}

Exact-match filtering does not detect paraphrases or contamination in pretraining data.

\begin{table}[H]
  \caption{Training samples removed by exact matching against each benchmark.}
  \label{tab:data-decontamination}
  \FusionEvalTableStyle
  \begin{tabular}{@{}l*{8}{c}@{}}
    \toprule
    \textbf{Domain} & \multicolumn{2}{c}{\textbf{Math}} & \multicolumn{2}{c}{\textbf{Code}} & \textbf{Science} & \multicolumn{2}{c}{\textbf{IF}} & \textbf{Tool use} \\
    \cmidrule(lr){2-3}\cmidrule(lr){4-5}\cmidrule(lr){6-6}\cmidrule(lr){7-8}\cmidrule(l){9-9}
    \textbf{Benchmark} & AIME24 & AIME25 & LCBv5 & LCBv6 & GPQA-D & IFEval & IFBench & BFCLv3 \\
    \midrule
    \rowcolor{black!5}
    \textbf{Removed} & $0$ & $0$ & $0$ & $0$ & $0$ & $0$ & $0$ & $1$ \\
    \bottomrule
  \end{tabular}
\end{table}

\subsection{Evaluation Protocols}
\label{app:evaluation-protocols}

Quick evaluation uses one generation per question and three BFCLv3 questions from each of 17 subsets.

All candidates use the same preset prefixes of fixed dataset snapshots.
Both stages use the same snapshots, so full evaluation is not an independent held-out test.

\begin{table}[H]
  \caption{Sample counts and decoding settings for quick evaluation.}
  \label{tab:evaluation-protocols}
  \FusionEvalTableStyle
  \begin{tabular}{@{}lccc@{}}
    \toprule
    \textbf{Benchmark} & \textbf{Samples} & \textbf{Temperature} & \textbf{Max. output tokens} \\
    \midrule
    AIME24 & 30 & 0.7 & 32768 \\
    AIME25 & 30 & 0.7 & 32768 \\
    LCBv5 & 20 & 0.2 & 32768 \\
    IFEval & 20 & 0.7 & 8192 \\
    BFCLv3 & 51 & 0.6 & 32768 \\
    \bottomrule
  \end{tabular}
\end{table}

\begin{table}[H]
  \caption{Decoding settings for full evaluation.}
  \label{tab:evaluation-decoding}
  \FusionEvalTableStyle
  \begin{tabular}{@{}lcc@{}}
    \toprule
    \textbf{Benchmark} & \textbf{Temp. / top-$p$} & \textbf{Prompt} \\
    \midrule
    AIME24 / AIME25 & 0.7 / 0.8 & Qwen \\
    LCBv5 / LCBv6 & 0.2 / 0.95 & Neutral \\
    GPQA-D & 0.7 / 0.8 & Qwen \\
    IFEval & 0.7 / 0.8 & Qwen \\
    IFBench & 0.0 / 1.0 & Qwen \\
    BFCLv3 & 0.6 / 0.95 & Qwen + tools \\
    \bottomrule
  \end{tabular}
\end{table}

Quick and full evaluations use seed 42 and top-k of 20.

\subsection{Efficiency Measurement Details}
\label{app:data-access-efficiency}

Runtime and peak GPU memory are measured on H800 GPUs. Parameter-fusion timings exclude weight loading, output saving, and checks, except that the recorded 8B RegMean run includes input reading during statistic construction. For repeated runs, we report mean runtime with min--max ranges and the largest memory peak in GiB. Some timings use configurations other than the selected evaluation checkpoints.

Distillation runtime sums the recorded rollout and training stages for MOPD, and teacher generation and student training for Seq-KD, excluding identifiable initialization, I/O, and idle intervals. Concurrent generation intervals are counted once. Both methods have a planned budget of 50 updates. MOPD uses 16 GPUs; Seq-KD uses 8 GPUs for training. Runtime covers the updates completed in the recorded runs.

\subsection{Hyperparameter Tuning}
\label{app:hyperparameter-tuning}

Candidate counts exclude full evaluations and internal optimization steps; $0$ denotes a fixed configuration.

Table~\ref{tab:hyperparameter-search} lists the search spaces and numbers of candidate validations.
These follow the original methods, MergeBench~\citep{he2025mergebench}, and the configurations used in our experiments.
We estimate search cost as mean Algorithm runtime multiplied by the validation count.
This estimate does not replace timing a complete search.

\begin{table}[!htbp]
  \caption{Fusion hyperparameter search spaces and candidate counts.}
  \label{tab:hyperparameter-search}
  \FusionEvalTableStyle
  \begin{tabularx}{\linewidth}{@{}>{\raggedright\arraybackslash}p{0.20\linewidth} c >{\raggedright\arraybackslash}X@{}}
    \toprule
    \textbf{Method} & \textbf{Candidates} & \textbf{Search space} \\
    \midrule
    Soup & $0$ & Uniform average \\
    Fisher Merging & $6$ & Number of samples per example $N\in\{4,8\}$; damping floor $\{10^{-7},10^{-6}\}$; target-model favoring; Fisher normalization \\
    AdaMerging & $3$ & Learning rate $\{3\times10^{-4},10^{-3}\}$; initialization $\{0.2,0.3\}$; embedding coefficient $\{\mathrm{mean},\mathrm{learned}\}$; $500$ optimization steps \\
    TA & $10$ & Scaling coefficient $\lambda\in\{0.1,0.2,\ldots,1.0\}$ \\

    LOT-Merging & $5$ & Scaling coefficient $\lambda_{\mathrm{LOT}}\in\{0.4,0.6,0.8,1.0,1.2,1.4\}$; $128$ calibration samples per domain \\
    TIES & $30$ & Retained fraction $\rho\in\{0.1,0.2,0.3\}$; scaling coefficient $\lambda\in\{0.1,0.2,\ldots,1.0\}$ \\
    DARE & $30$ & Drop rate $p\in\{0.1,0.2,0.3\}$; scaling coefficient $\lambda\in\{0.1,0.2,\ldots,1.0\}$ \\
    DELLA & $30$ & Drop rate $p\in\{0.1,0.2,0.3\}$; scaling $\lambda\in\{0.1,0.2,\ldots,1.0\}$; window fixed to $0.14$ \\
    L\&S & $0$ & Retained fraction fixed to $\rho=0.40$ \\
    RegMean & $6$ & Shrinkage coefficient $\alpha\in\{0.1,0.3,0.5,0.7,0.9,1.0\}$; $128$ calibration samples per domain \\
    \bottomrule
  \end{tabularx}
\end{table}

\subsection{Selected Fusion Configurations}
\label{app:selected-fusion-configurations}

Tables~\ref{tab:selected-config-q4}--\ref{tab:selected-config-llama} give the configurations corresponding to the reported models in each expert pool. $N$ denotes predictive samples for Fisher estimation. Data requirements are listed in Table~\ref{tab:merging-methods}. Calibration uses 128 examples per domain; distillation uses a constant learning rate.

\FloatBarrier
\begin{table}[p]
  \FusionEvalTableStyle
  \caption{Selected fusion configurations for Qwen3-4B.}
  \label{tab:selected-config-q4}\label{tab:selected-fusion-configurations}
  \begin{tabularx}{\linewidth}{@{}l >{\raggedright\arraybackslash}X@{}}
    \toprule
    \textbf{Method} & \textbf{Configuration} \\
    \midrule
    Soup & Uniform average ($1/5$ per expert) \\
    TA & $\lambda=0.3$ \\
    TIES & Density $0.2$; scale $1.0$; disjoint mean \\
    DARE & Drop $0.2$; scale $0.3$; seed 42 \\
    DELLA & Drop $0.1$; window $0.14$; scale $1.0$; seed 42 \\
    L\&S & Retained fraction $0.40$ \\
    Fisher & Coefficients $0.2$; predictive Fisher, $N=8$; normalized; floor $10^{-6}$ (all experts); full prompts \\
    Ada & Init.\ $0.3$; lr $3\times10^{-4}$; $500$ steps; embedding mean; prompt cap $4096$ \\
    LOT & scale $1.0$; closed-form Linear/RMSNorm \\
    RegMean & $\alpha=0.7$; prefix $4096$; raw Gram, FP32; mean for tied embedding/head and 1D weights \\
    Soup+Surgery & Rank $32$; lr $10^{-4}$; $250$ steps \\
    Soup+FeatCal & $\lambda=0.01$; $\rho=1.5$; $\alpha=0.25$ \\
    TA+Surgery & Rank $16$; lr $6\times10^{-5}$; $250$ steps \\
    TA+FeatCal & $\lambda=1$; $\rho=1.5$; $\alpha=0.25$ \\
    Seq-KD & 50 updates; batch $512$; lr $3\times10^{-6}$; response-token loss; prefix $4096$ \\
    MOPD & 30 updates; batch $512$; lr $3\times10^{-6}$; Adam; temperature $1.0$; top-$p$ $1.0$; output cap $32768$ \\
    \bottomrule
  \end{tabularx}

  \medskip

  \caption{Selected fusion configurations for Qwen3-8B.}
  \label{tab:selected-config-q8}
  \begin{tabularx}{\linewidth}{@{}l >{\raggedright\arraybackslash}X@{}}
    \toprule
    \textbf{Method} & \textbf{Configuration} \\
    \midrule
    Soup & Uniform average ($1/5$ per expert) \\
    TA & $\lambda=0.3$ \\
    TIES & Density $0.2$; scale $0.9$; disjoint mean \\
    DARE & Drop $0.1$; scale $0.3$; seed 42 \\
    DELLA & Drop $0.1$; window $0.14$; scale $1.0$; seed 42 \\
    L\&S & Retained fraction $0.40$ \\
    Fisher & Coefficients $0.2$; predictive Fisher, $N=4$; normalized; floor $10^{-6}$ (first expert); prefix $4096$ \\
    Ada & Init.\ $0.3$; lr $3\times10^{-4}$; $500$ steps; embedding mean; prompt cap $4096$ \\
    LOT & scale $1.4$; closed-form Linear/RMSNorm \\
    RegMean & $\alpha=0.7$; prefix $4096$; raw Gram, FP32; mean for embedding and 1D weights; RegMean head \\
    Soup+Surgery & Rank $8$; lr $10^{-4}$; $250$ steps \\
    Soup+FeatCal & $\lambda=0.1$; $\rho=1.5$; $\alpha=0.15$ \\
    TA+Surgery & Rank $32$; lr $10^{-4}$; $500$ steps \\
    TA+FeatCal & $\lambda=0.1$; $\rho=1.5$; $\alpha=0.1$ \\
    Seq-KD & 50 updates; batch $512$; lr $3\times10^{-6}$; response-token loss; prefix $4096$ \\
    MOPD & 40 updates; batch $512$; lr $3\times10^{-6}$; Adam; temperature $1.0$; top-$p$ $1.0$; output cap $32768$ \\
    \bottomrule
  \end{tabularx}
\end{table}

\FloatBarrier
\begingroup
\FusionEvalTableStyle
\setlength{\LTcapwidth}{\linewidth}
\begin{longtable}{@{}l >{\raggedright\arraybackslash}p{0.78\linewidth}@{}}
  \caption{Selected fusion configurations for Llama-3.2-3B.}\label{tab:selected-config-llama} \\
  \toprule
  \textbf{Method} & \textbf{Configuration} \\
  \midrule
  \endfirsthead
  \multicolumn{2}{c}{Table \thetable\ (continued)} \\
  \toprule
  \textbf{Method} & \textbf{Configuration} \\
  \midrule
  \endhead
    TA & $\lambda=0.3$ \\
    TIES & Density $0.2$; scale $1.0$; disjoint mean \\
    DARE & Drop $0.2$; scale $0.3$ \\
    DELLA & Drop $0.1$; window $0.14$; scale $1.0$ \\
    TA+FeatCal & $\lambda=0.1$; $\rho=1.5$; $\alpha=0.25$; $128$/domain \\
    Consensus TA & TALL threshold $0.2$; minimum tasks $2$; scale $0.3$ \\
      \bottomrule
\end{longtable}
\endgroup

\subsection{Update Geometry Details}
\label{app:update-geometry}
\label{sec:parameter-space-analysis}
\label{app:tool-degradation}

\paragraph{Metrics and candidates.}
We compare Qwen3-4B backbone updates relative to the shared base, excluding embeddings and output heads. Update magnitude is the $L_2$ norm; Agent similarity is the cosine between fused and Agent updates. Let $\Delta$ and $\Delta_A$ denote these updates. Agent-weighted sign conflict is

\begin{equation}
 C_A=\frac{\sum_j |\Delta_{A,j}|\,\mathbf{1}[\Delta_j\Delta_{A,j}<0]}{\sum_j |\Delta_{A,j}|}.
 \label{eq:agent-sign-conflict}
\end{equation}
where $j$ indexes parameters. The zero-update fraction uses the same denominator with the condition $\Delta_j=0$.

For the candidates in Figures~\ref{fig:tool-degradation} and~\ref{fig:tool-tradeoff}, F1--F4 use Fisher floors $10^{-6},10^{-7},10^{-8},10^{-10}$ with fixed statistics (128 calibration examples per domain, $N=8$). D1--D3 use DARE (drop, scale) values $(0.2,0.3)$, $(0.2,0.4)$, and $(0.1,0.2)$; T uses TIES density $0.3$ and scale $1$. S denotes Soup, and A denotes TA with scale $0.3$. Scores use the quick-evaluation samples in Table~\ref{tab:evaluation-protocols}; BFCLv3 screening scores use hierarchical aggregation across subsets.

\begin{table}[H]
  \FusionEvalTableStyle
  \caption{Agent-weighted sign conflict and zero-update fractions on Qwen3-4B. Candidate labels follow the configurations defined above.}
  \label{tab:agent-sign-conflict}
  \begin{tabular}{@{}lrr@{\hspace{18pt}}lrr@{}}
    \toprule
    \textbf{Model} & \textbf{Conflict (\%)} & \textbf{Zero (\%)} & \textbf{Model} & \textbf{Conflict (\%)} & \textbf{Zero (\%)} \\
    \midrule
    F1 & 42.50 & 1.93 & D1 & 32.85 & 1.89 \\
    F2 & 36.93 & 2.11 & D2 & 33.12 & 1.32 \\
    F3 & 26.36 & 2.20 & D3 & 31.88 & 2.86 \\
    F4 & 20.73 & 2.01 & S & 31.54 & 2.69 \\
    T & 28.85 & 9.74 & A & 32.02 & 1.69 \\
    \bottomrule
  \end{tabular}
\end{table}

\paragraph{Fisher response examples.}
\label{app:fisher-response-cases}
Table~\ref{tab:fisher-response-cases} compares two questions across higher-, middle-, and lower-scoring Fisher candidates (H/M/L), with full BFCLv3 scores of $57.9/45.1/29.9$. 

\begin{table}[H]
  \FusionEvalTableStyle
  \caption{Fisher response summaries and evaluation outcomes on two BFCLv3 questions.}
  \label{tab:fisher-response-cases}
  \begin{tabularx}{\linewidth}{@{}lXl@{}}
    \toprule
    \textbf{Model} & \textbf{Response summary} & \textbf{Result} \\
    \midrule
    \multicolumn{3}{@{}l}{\textbf{\texttt{multiple\_0}}\quad Triangle properties for sides $5,4,3$} \\
    H & Structured call to \texttt{triangle\_properties\_get} with \texttt{side1=5, side2=4, side3=3}. & Pass \\
    M & The same structured function call and arguments as H. & Pass \\
    L & Identifies the correct function and sides in text, but discusses XML wrappers without producing a decodable call. & Fail \\
    \midrule
    \multicolumn{3}{@{}l}{\textbf{\texttt{multiple\_100}}\quad Highest common factor of $36$ and $24$} \\
    H & Structured call to \texttt{math\_hcf} with \texttt{number1=36, number2=24}. & Pass \\
    M & Identifies the correct function and arguments, but switches between wrapper formats; the output is parsed as text. & Fail \\
    L & Identifies the correct function and arguments and computes $12$, but uses a \texttt{<tools>} wrapper; the output is parsed as text. & Fail \\
    \bottomrule
  \end{tabularx}
\end{table}

\subsection{Additional Fusion Results}
\label{app:additional-fusion-results}

\begin{table}[!htbp]
  \caption{Additional fusion and calibration results on Qwen3. Scores follow Tables~\ref{tab:qwen3-4b-fusion-results} and~\ref{tab:parameter-level-8b-results}.}
  \label{tab:additional-fusion-results}
  \FusionEvalTableStyle
  \setlength{\tabcolsep}{1.6pt}
  \begin{tabular}{@{}l*{11}{c}@{}}
    \toprule
    & \multicolumn{2}{c}{\textbf{Math}} & \multicolumn{2}{c}{\textbf{Code}} & \textbf{Science} & \multicolumn{2}{c}{\textbf{IF}} & \textbf{Tool use} & \textbf{Raw} & \multicolumn{2}{c}{\textbf{Norm.}} \\
    \cmidrule(lr){2-3}\cmidrule(lr){4-5}\cmidrule(lr){6-6}\cmidrule(lr){7-8}\cmidrule(lr){9-9}\cmidrule(l){11-12}
\textbf{Method} & \textbf{AIME24} & \textbf{AIME25} & \textbf{LCBv5} & \textbf{LCBv6} & \textbf{GPQA-D} & \textbf{IFEval} & \textbf{IFBench} & \textbf{BFCLv3} & \textbf{Avg.} & \textbf{Avg.} & \textbf{Min.} \\
    \midrule
    \multicolumn{12}{@{}l}{\textbf{Qwen3-4B}} \\
    LOT & 39.4 & 35.6 & 23.7 & 26.2 & 42.3 & 65.1 & 26.2 & 30.8 & 36.1 & 0.618 & 0.466 \\
    Soup+FeatCal & 39.4 & 34.8 & 31.7 & 28.2 & 42.3 & 70.8 & 33.4 & 31.2 & 39.0 & 0.678 & 0.473 \\
    Soup+Surgery & 43.3 & 32.1 & 28.9 & 28.6 & 44.8 & 65.5 & 29.3 & 31.8 & 38.1 & 0.659 & 0.476 \\
    TA+Surgery & 66.0 & 56.9 & 45.3 & 45.2 & 46.1 & 73.2 & 39.2 & 31.4 & 50.4 & 0.885 & 0.475 \\
    TA+FeatCal & 68.8 & 67.5 & 46.6 & 39.9 & 38.3 & 66.6 & 39.1 & 31.9 & 49.8 & 0.869 & 0.483 \\
    \midrule
    \multicolumn{12}{@{}l}{\textbf{Qwen3-8B}} \\
    LOT & 51.3 & 49.0 & 33.9 & 30.5 & 42.4 & 78.7 & 36.8 & 29.8 & 44.0 & 0.735 & 0.441 \\
    Soup+FeatCal & 58.1 & 51.9 & 49.5 & 41.0 & 54.4 & 77.3 & 41.7 & 30.2 & 50.5 & 0.865 & 0.448 \\
    Soup+Surgery & 50.4 & 36.9 & 36.6 & 34.4 & 56.7 & 76.6 & 33.4 & 33.0 & 44.7 & 0.756 & 0.489 \\
    TA+FeatCal & 70.4 & 66.7 & 60.2 & 49.6 & 50.0 & 77.5 & 42.2 & 31.4 & 56.0 & 0.959 & 0.465 \\
    TA+Surgery & 64.4 & 60.0 & 54.7 & 44.5 & 45.8 & 76.3 & 39.7 & 34.3 & 52.5 & 0.893 & 0.508 \\
    \bottomrule
  \end{tabular}
\end{table}

\subsection{Continual Fusion Results}
\label{app:continual-results}

Stage 0 is the IF expert; Stages 1--4 add Math, Science, Code, and Agent. 
\paragraph{Update rules.}
Continual Soup assigns equal weight to the current model and the incoming expert at each step.
TA starts from the same IF expert and accumulates task vectors relative to the fixed base. Let $\hat\theta_t$ be the model after stage $t$ and $\pi(t)$ the incoming expert:
\begin{equation}
  \label{eq:continual-updates}
  \begin{aligned}
    \hat\theta_t^{\mathrm{Avg}}
      &=\tfrac12\hat\theta_{t-1}^{\mathrm{Avg}}+\tfrac12\theta_{\pi(t)},\\
    \hat\theta_1^{\mathrm{TA}}
      &=\theta_0+\lambda(\theta_{\mathrm{IF}}-\theta_0)+\lambda(\theta_{\pi(1)}-\theta_0),\\
    \hat\theta_t^{\mathrm{TA}}
      &=\hat\theta_{t-1}^{\mathrm{TA}}+\lambda(\theta_{\pi(t)}-\theta_0),
      \qquad t\geq2,\quad\lambda=0.3.
  \end{aligned}
\end{equation}

\begin{table}[H]
  \caption{Per-benchmark scores (\%) at each Continual Fusion stage on Qwen3-4B.}
  \label{tab:continual-final-scores}
  \FusionEvalTableStyle
  \setlength{\tabcolsep}{1.6pt}
  \begin{tabular}{@{}lc*{8}{c}@{}}
    \toprule
    \textbf{Method} & \textbf{Stage} & \textbf{AIME24} & \textbf{AIME25} & \textbf{LCBv5} & \textbf{LCBv6} & \textbf{GPQA-D} & \textbf{IFEval} & \textbf{IFBench} & \textbf{BFCLv3} \\
    \midrule
    IF expert & 0 & 16.7 & 17.1 & 14.2 & 14.1 & 20.2 & 86.0 & 38.6 & 29.8 \\
    \midrule
    \multirow{4}{*}{TA} & 1 & 24.6 & 20.4 & 14.7 & 17.6 & 37.4 & 51.0 & 21.1 & 29.7 \\
     & 2 & 42.5 & 27.1 & 23.7 & 21.4 & 46.0 & 69.5 & 29.1 & 29.7 \\
     & 3 & 62.9 & 56.4 & 37.3 & 37.8 & 43.4 & 72.6 & 36.1 & 29.7 \\
     & 4 & 65.4 & 65.8 & 45.9 & 42.7 & 45.1 & 71.0 & 38.0 & 30.9 \\
    \midrule
    \multirow{4}{*}{Continual Soup} & 1 & 42.3 & 36.5 & 20.8 & 23.3 & 41.4 & 69.7 & 30.5 & 29.8 \\
     & 2 & 50.0 & 41.5 & 22.9 & 25.2 & 49.6 & 66.9 & 28.6 & 29.7 \\
     & 3 & 53.3 & 45.8 & 42.1 & 35.3 & 41.0 & 62.4 & 30.3 & 29.8 \\
     & 4 & 25.8 & 26.3 & 26.7 & 26.5 & 48.5 & 59.4 & 23.2 & 30.0 \\
    \midrule
    \multirow{4}{*}{Continual TIES} & 1 & 26.7 & 27.9 & 2.5 & 3.8 & 11.1 & 28.8 & 17.8 & 29.7 \\
     & 2 & 57.9 & 46.7 & 7.9 & 9.2 & 34.1 & 41.8 & 20.2 & 29.7 \\
     & 3 & 38.8 & 35.4 & 7.2 & 9.5 & 15.7 & 30.9 & 23.7 & 29.7 \\
     & 4 & 41.0 & 33.3 & 27.1 & 27.7 & 42.2 & 53.5 & 23.2 & 50.7 \\
    \midrule
    \multirow{4}{*}{Continual DELLA} & 1 & 65.2 & 52.1 & 27.2 & 31.7 & 33.1 & 62.7 & 30.3 & 29.8 \\
     & 2 & 66.7 & 54.8 & 30.8 & 27.9 & 47.2 & 59.9 & 27.3 & 29.6 \\
     & 3 & 67.5 & 64.2 & 49.1 & 40.8 & 45.5 & 61.6 & 33.3 & 29.8 \\
     & 4 & 58.8 & 47.3 & 42.7 & 36.8 & 48.6 & 58.6 & 30.2 & 30.3 \\
    \bottomrule
  \end{tabular}
\end{table}

\section{Distillation Data}
\label{app:distillation-data}

\paragraph{Prompt sources.}
Table~\ref{tab:mopd-data} lists the distillation prompt pool and counts after preprocessing; experiment-specific subsets are selected separately. MOPD loads prompts with an $8{,}192$-token limit. Agent prompts retain tool definitions and dialogue history, without executing tool calls during training.
Seq-KD uses the fixed $25{,}600$ prompts scheduled for the first $50$ steps in the original MOPD data-loading and shuffle order. Both model sizes use the same prompts, with responses generated by the corresponding 4B or 8B domain experts. 
\begin{table}[H]
  \caption{Distillation prompt sources and counts after preprocessing.}
  \label{tab:mopd-data}
  \FusionEvalTableStyle
  \begin{tabularx}{\linewidth}{@{}l r >{\raggedright\arraybackslash}X@{}}
    \toprule
    \textbf{Domain} & \textbf{Prompts} & \textbf{Sources (prompts per source)} \\
    \midrule
    Math & 18,618 & Skywork/Skywork-OR1-RL-Data (12,157); nvidia/Nemotron-RL-Math-v2 (6,461) \\
    Code & 19,954 & nvidia/Nemotron-RL-coding-competitive\_coding (11,997); IIGroup/X-Coder-RL-40k (7,957) \\
    Science & 19,960 & nvidia/Nemotron-RL-knowledge-mcqa (19,960) \\
    IF & 19,758 & allenai/IF\_multi\_constraints\_upto5 (11,861); nvidia/Nemotron-RL-instruction\_following (7,897) \\
    Agent & 8,735 & NVIDIA Function-Calling Pivot (7,752); nvidia/Nemotron-Cascade-2-RL-data (983) \\
    \bottomrule
  \end{tabularx}
\end{table}

\end{document}

%% file: math_commands.tex
\usepackage{amsmath,amsfonts,bm}

\def\eqref#1{equation~\ref{#1}}
\def\1{\bm{1}}

\DeclareMathAlphabet{\mathsfit}{\encodingdefault}{\sfdefault}{m}{sl}
\SetMathAlphabet{\mathsfit}{bold}{\encodingdefault}{\sfdefault}{bx}{n}